\documentclass[a4paper,fleqn]{cas-dc}

\usepackage[numbers]{natbib}

\usepackage{hyperref}       
\usepackage{url}            
\usepackage{booktabs}       
\usepackage{amsfonts}       
\usepackage{nicefrac}       
\usepackage{microtype}      
\usepackage{xcolor}         
\usepackage{multirow}
\usepackage{graphicx}
\usepackage{bbding}
\usepackage{makecell}
\usepackage{amsthm,amsmath,amssymb}
\usepackage{adjustbox}
\usepackage{pifont}
\usepackage{subfigure}
\usepackage{tcolorbox}
\usepackage{parskip}
\definecolor{my_green}{RGB}{51,102,0}
\definecolor{my_red}{RGB}{204, 0, 0}
\renewcommand{\checkmark}{\textcolor{my_green}{\ding{51}}}
\newcommand{\crossmark}{\textcolor{my_red}{\ding{55}}}

\def\tsc#1{\csdef{#1}{\textsc{\lowercase{#1}}\xspace}}
\tsc{WGM}
\tsc{QE}
\tsc{EP}
\tsc{PMS}
\tsc{BEC}
\tsc{DE}

\begin{document}

\begin{sloppypar}
\let\WriteBookmarks\relax
\def\floatpagepagefraction{1}
\def\textpagefraction{.001}
\shorttitle{AIGCBench}
\shortauthors{Fanda Fan, Jianfeng Zhan}

\title [mode = title]{AIGCBench: Comprehensive Evaluation of Image-to-Video Content Generated by AI}

\author[1,2]{Fanda Fan}[type=editor,
                        auid=000,bioid=1,
                        orcid=0000-0002-5214-0959]
\ead{fanfanda@ict.ac.cn}

\author[1]{Chunjie Luo}[type=editor,
                        auid=000,bioid=1,
                        orcid=0000-0002-6977-929X]

\author[1]{Wanling Gao}[type=editor,
                        auid=000,bioid=1,
                        orcid=0000-0002-3911-9389]

\author[1,2]{Jianfeng Zhan}[type=editor,
                        auid=000,bioid=1,
                        orcid=0000-0002-3728-6837]
\cormark[1]
\ead{zhanjianfeng@ict.ac.cn}

\cortext[cor1]{Corresponding author}

\address[1]{Research Center for Advanced Computer Systems, State Key Lab of Processors, Institute of Computing Technology, Chinese Academy of Sciences, China}
\address[2]{University of Chinese Academy of Sciences, China}

\begin{abstract}
The burgeoning field of Artificial Intelligence Generated Content (AIGC) is witnessing rapid advancements, particularly in video generation. This paper introduces AIGCBench, a pioneering comprehensive and scalable benchmark designed to evaluate a variety of video generation tasks, with a primary focus on Image-to-Video (I2V) generation. AIGCBench tackles the limitations of existing benchmarks, which suffer from a lack of diverse datasets, by including a varied and open-domain image-text dataset that evaluates different state-of-the-art algorithms under equivalent conditions. We employ a novel text combiner and GPT-4 to create rich text prompts, which are then used to generate images via advanced Text-to-Image models. To establish a unified evaluation framework for video generation tasks, our benchmark includes 11 metrics spanning four dimensions to assess algorithm performance. These dimensions are control-video alignment, motion effects, temporal consistency, and video quality. These metrics are both reference video-based and video-free, ensuring a comprehensive evaluation strategy. The evaluation standard proposed correlates well with human judgment, providing insights into the strengths and weaknesses of current I2V algorithms. The findings from our extensive experiments aim to stimulate further research and development in the I2V field. AIGCBench represents a significant step toward creating standardized benchmarks for the broader AIGC landscape, proposing an adaptable and equitable framework for future assessments of video generation tasks. We have open-sourced the dataset and evaluation code on the project website: \href{https://www.benchcouncil.org/AIGCBench}{https://www.benchcouncil.org/AIGCBench}.

\end{abstract}




\maketitle

\section{Introduction}

Artificial Intelligence Generated Content (AIGC) encompasses a wide array of applications that leverage AI technologies to automate the creation or editing of content across different media types, such as text, images, audio, and video. With the rapid advancement of diffusion models~\cite{sohl2015deep,song2019generative,ho2020denoising,nichol2021improved,dhariwal2021diffusion} and multimodal AI technologies~\cite{radford2021learning}, the AIGC field is experiencing considerable and rapid progress. The explosive growth of AIGC has made its evaluation and benchmarking an urgent task.

A representative application of AIGC is video generation~\cite{singer2022make,ho2022imagen,discordpika,esser2023structure,sun2023generative}. Current video generation includes Text-to-Video (T2V), Image-to-Video (I2V), Video-to-Video (V2V), as well as a few other works that utilize additional information such as depth~\cite{esser2023structure}, pose~\cite{karras2023dreampose}, trajectory~\cite{yin2023dragnuwa}, and frequency~\cite{li2023generative} to generate videos. Among these, T2V and I2V are the two most mainstream tasks at present. Early video generation primarily used text prompts to generate videos and achieved good results~\cite{hong2022cogvideo,singer2022make,ho2022imagen,he2022latent,wu2023tune,luo2023videofusion,guo2023animatediff}. However, using text alone makes it difficult to depict the specific scenes that users want. Recently, I2V has ignited the AIGC community. The I2V task refers to the generation of a dynamic, moving video sequence based on a static input image and is usually accompanied by a text prompt~\footnote{However, the community often refers to it as Image-to-Video, rather than Text-Image-to-Video.}. Compared to T2V, I2V can better define the content of video generation, achieving excellent results in many scenarios such as film, e-commerce advertising, and micro-animation effects.

While benchmarks for the T2V task have seen notable progress~\cite{liu2023fetv,liu2023evalcrafter,huang2023vbench}, benchmarks for the I2V task have scarcely advanced.
Previous efforts like Latent Flow Diffusion Models~(LFDM)~\cite{ni2023conditional} and CATER-GEN~\cite{hu2023benchmark} were tested under domain-specific video scenarios. VideoCrafter~\cite{chen2023videocrafter1} and I2VGen-XL~\cite{zhang2023i2vgen} only utilized visual comparisons for the I2V task. Seer~\cite{gu2023seer} and Stable Video Diffusion~(SVD)~\cite{blattmann2023stable} employed video-text datasets and utilized a few metrics that require reference videos. Existing I2V benchmarks suffer from 1)\label{sec:firstshort} a lack of diverse, open-domain images\footnote{Open-domain images refer to images that cover a wide variety of subjects or topics without specific restrictions on the content or category.} with various subjects and styles to test the efficacy of different state-of-the-art algorithms;
2) an absence of a unified consensus on which evaluation metrics should be used to assess the final generated results. From the perspective of~\cite{Zhan2023Evaluatology}, these two shortcomings hinder the capability of capturing stakeholders' concerns and interests, while also failing to construct equivalent evaluation conditions.

\begin{figure*}
    \centering
    \includegraphics[scale=.55]{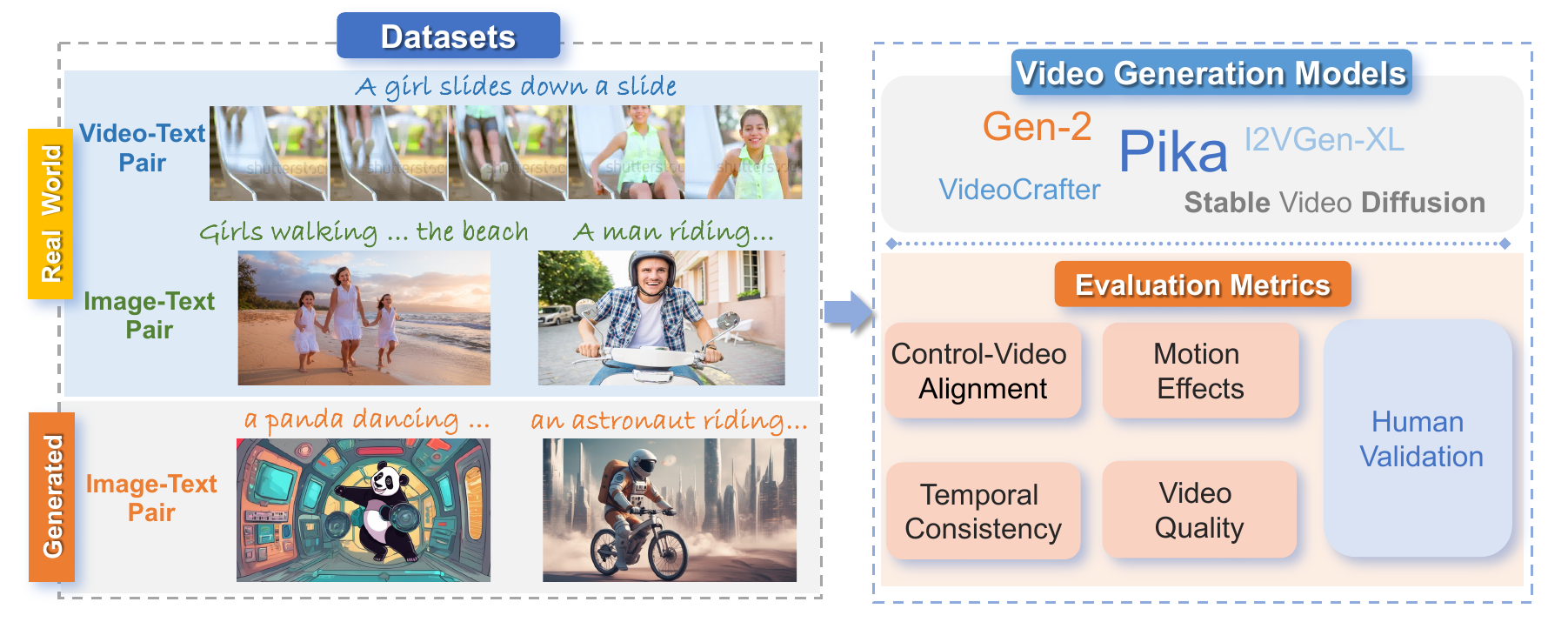}
    \caption{Illustration of our AIGCBench. Our AIGCBench is divided into three modules: the evaluation dataset, the evaluation metrics, and the video generation models to be assessed. Our benchmark encompasses two types of datasets: video-text and image-text datasets. To construct a more comprehensive evaluation dataset, we expand the image-text dataset by our generation pipeline. Additionally, for a thorough evaluation of video generation models, we introduce a set of evaluation metrics comprising 11 metrics across four dimensions. These metrics include both reference video-based and reference video-free metrics, making full use of the benchmark we propose. We also adopted human validation to confirm the rationality of the evaluation standards we proposed.}
    \label{fig:framework}
\end{figure*}


To address this gap, we present AIGCBench, a unified benchmark for video generation tasks. AIGCBench aims to encapsulate all mainstream video generation tasks, such as T2V, I2V, V2V, and the synthesis of video from additional modalities like depth, pose, trajectory, and frequency. We present an overview of AIGCBench in Figure~\ref{fig:framework}. Our AIGCBench is divided into three modules: the evaluation dataset, the evaluation metrics, and the video generation models to be assessed. Considering the high relevance and interconnectivity~\footnote{The interconnectivity arises because some algorithms have the capability to perform multiple types of video generation tasks.} of video generation tasks, our AIGCBench can enable the comparison of different algorithms under equivalent evaluation conditions. This allows for an analysis of the strengths and weaknesses of different state-of-the-art video generation algorithms, thereby aiding progress in the field of video generation. In the first version of our AIGCBench, we address the current lack of a reasonable benchmark for I2V tasks by providing a thorough evaluation for them. In subsequent versions, we plan to include more video generation tasks and place them under equivalent evaluation conditions for a fair comparison.



Recognizing the limitations of existing benchmarks, AIGCBench is engineered to meet the diverse demands of users looking to animate a broad array of static images. Where previous benchmarks have fallen short, not fully accommodating the expansive range of images users might choose to animate — such as a blue dragon skateboarding in Times Square — AIGCBench rises to the challenge. We address this by deploying a text combiner to generate a rich assortment of text prompts that span a multitude of subjects, behaviors, backgrounds, and artistic styles. Further refining the creative process, we employ the advanced capabilities of GPT-4~\cite{openai2023gpt4} to enhance the text prompts, rendering them more vivid and intricate. These detailed prompts then guide the generation of images through state-of-the-art Text-to-Image diffusion models. By judiciously blending video-text and image-text datasets, along with our generated image-text pairs, AIGCBench ensures a robust and comprehensive evaluation of an array of I2V algorithms, thus addressing the first major shortcoming identified in existing benchmarks. 



To establish a comprehensive and standardized set of evaluation metrics for video generation tasks that cater to mainstream tasks such as T2V and I2V,  our AIGCBench evaluates four critical dimensions: control-video alignment, motion effects, temporal consistency, and video quality, thereby capturing every aspect of video generation. This integrated framework combines metrics that are both reference video-based and video-free metrics, enhancing the benchmark's rigor without exclusively relying on video-text datasets or image-text datasets alone. We strengthen this approach by incorporating image-text datasets into our evaluations, which allows us to assess content beyond the scope of existing video-text datasets and add reference video-free metrics for assessment. Considering the complexity and diversity of tasks, we believe that the evaluation metrics should cover at least these four aspects. For each aspect, we aim to use both reference video-based and video-free metrics. After satisfying these categorizations, the benefits of increasing the number of metrics become marginal, while it is insufficient without covering these aspects. The experimental results demonstrate that our evaluation standard correlates well with human ratings, confirming its effectiveness. Following a thorough evaluation, we present the strengths and weaknesses of each model, alongside several insightful findings, in hopes of spurring discussions that advance the I2V field.

Our contributions are as follows:

\begin{enumerate}
\itemsep=0pt
\item We introduce AIGCBench, a benchmark for comprehensive evaluation of diverse video generation tasks, with an initial focus on Image-to-Video (I2V) generation and a commitment to placing these models under equivalent evaluation conditions for fair comparison.
\item We extend our image-text dataset using a text combiner and GPT-4, complemented by state-of-the-art Text-to-Image models to generate high-quality images, enabling a deeper evaluation of I2V algorithm performance;
\item We evaluate I2V algorithms comprehensively using both reference video-based and video-free metrics across four aspects and verify the validity of our proposed evaluation standard with human judgment;
\item We offer several insightful findings to aid the better development of the I2V community.

\end{enumerate}  


\begin{table*}[t]
\centering
\caption{Compare the features of our AIGCBench with those of others. \crossmark ~ and \checkmark indicate whether the benchmark includes the features listed in the respective columns. Video-based metrics, which use reference videos, contrast with video-free metrics that don't. Considering the difficulty of the evaluation, we are not counting the sample numbers for domain-specific benchmarks~\cite{ni2023conditional,hu2023benchmark,gu2023seer}.}
\label{tab:compare_benchmark}
\resizebox{\textwidth}{!}{
\begin{tabular}{lccccccccc}
\toprule
Benchmark  & Open-Domain & Video-Text Pairs & Image-Text Pairs & Generated Dataset & \#Samples & Metric Types & \# Metrics \\  \midrule
LFDM Eval~\cite{ni2023conditional} & \crossmark & \checkmark & \crossmark & \crossmark & - & Video-based & 3 \\
CATER-GEN~\cite{hu2023benchmark} & \crossmark & \checkmark & \checkmark & \checkmark & - & Video-based \& Video-free & 7 \\
Seer Eval~\cite{gu2023seer} & \crossmark & \checkmark & \crossmark & \crossmark & - & Video-based & 2 \\
VideoCrafter Eval~\cite{chen2023videocrafter1} & \checkmark & \checkmark & \checkmark & \crossmark & - & - & - \\
I2VGen-XL Eval~\cite{zhang2023i2vgen} & \checkmark & \checkmark & \checkmark & \crossmark & - & - & - \\
SVD Eval~\cite{blattmann2023stable} & \checkmark & \checkmark & \checkmark & \crossmark & 900 & Video-based & 5 \\
AnimateBench~\cite{zhang2023pia} & \checkmark & \crossmark & \crossmark & \checkmark & 105 & Video-free & 2 \\
\rowcolor{gray! 40} AIGCBench (Ours) & \checkmark & \checkmark & \checkmark & \checkmark & 3928 & Video-based \& Video-free & 11 \\ \bottomrule
\end{tabular}
}
\end{table*}


\section{Background and Related Work}
\label{sec:related_work}


Current video generation primarily encompasses two major tasks: Text-to-Video (T2V) and Image-to-Video (I2V). Given the high relevance of T2V tasks to I2V tasks, we discuss video generation models, with a particular focus on I2V models, in Section~\ref{sec:i2vhistory}. We will introduce related benchmarks for T2V in Section~\ref{sec:t2vbench} and describe the existing benchmarks for I2V in Section~\ref{sec:i2vbench_related}.


\subsection{Video Generation Models}
\label{sec:i2vhistory}

Thanks to the development of diffusion models~\cite{sohl2015deep,song2019generative,ho2020denoising,nichol2021improved,dhariwal2021diffusion} and multimodal techniques~\cite{radford2021learning}, video generation algorithms are becoming increasingly sophisticated. Early video generation was primarily based on text-to-video approaches~\cite{hong2022cogvideo,singer2022make,ho2022imagen,wu2023tune,blattmann2023align,yu2023magvit,luo2023videofusion,ge2023preserve,khachatryan2023text2video,esser2023structure,wang2023modelscope,guo2023animatediff}. Most of the work is based on diffusion models~\cite{song2019generative,ho2020denoising,nichol2021improved,dhariwal2021diffusion,he2022latent}, with some being transformer-based~\cite{hong2022cogvideo,yu2023magvit}. They all rely on extensive video-text or image-text datasets to train scalable models. However, considering that using only text can make it challenging to intuitively depict the video scenes users want to generate, image-to-video has started to gain popularity in the video generation community.

Seer~\cite{gu2023seer} introduced an approach for I2V tasks that combines the conditional image latent with a noisy latent, utilizing causal attention within the temporal component of a 3D U-Net~\cite{ronneberger2015u}. 
VideoComposer~\cite{wang2023videocomposer} concatenated image embedding with image style embedding to preserve the initial image information. 
Recently, VideoCrafter~\cite{chen2023videocrafter1} encoded the image prompt through a lightweight image encoder and fed it into the cross-attention layer.
Similarly, I2VGen-XL~\cite{zhang2023i2vgen} not only merges the image latent with the noisy latent at the input layer but also employs a global encoder that extracts the image CLIP feature into the video latent diffusion model (VLDM).
Stable video diffusion~\cite{blattmann2023stable} is an extension of a pretrained image-based diffusion model~\cite{rombach2022high}. It is trained through three stages: text-to-image pretraining, video pretraining, and high-quality video fine-tuning.
Emu Video~\cite{girdhar2023emu} identified critical design decisions, such as adjusted noise schedules for diffusion and multi-stage training, which enabled the generation of high-quality videos without requiring a deep cascade of models as in prior work.
Beyond academic research, the video generation results from industry players like Pika~\cite{discordpika} and Gen2~\cite{esser2023structure} are also quite impressive. All of these I2V algorithms are based on video diffusion models, and the majority leverage the parameter priors from image diffusion models to aid in the convergence of video models. 

To evaluate state-of-the-art I2V models, we have reviewed three open-source works in this paper: VideoCrafter~\cite{chen2023videocrafter1}, I2VGen-XL~\cite{zhang2023i2vgen}, and Stable Video Diffusion~\cite{blattmann2023stable}, as well as two closed-source industry efforts, Pika~\cite{discordpika} and Gen2~\cite{esser2023structure}. These currently represent the five most influential works in the video generation community, and we will briefly introduce their experimental parameters in Section~\ref{sec:exp_models}.

\subsection{Benchmarks for Text-to-Video Generation}
\label{sec:t2vbench}
The FETV benchmark~\cite{liu2023fetv} conducts a comprehensive manual evaluation of representative T2V models and reveals their strengths and weaknesses in handling a diverse range of text prompts from multiple perspectives. 
EvalCrafter~\cite{liu2023evalcrafter} starts by creating a new set of prompts for T2V generation with the assistance of a large language model, ensuring that the prompts are representative of actual user queries. 
EvalCrafter's benchmarks~\cite{liu2023evalcrafter} are meticulously designed to evaluate generated videos from several critical dimensions: visual quality, content accuracy, motion dynamics, and the alignment between generated video content and the original text captions.
VBench~\cite{huang2023vbench} has created 16 distinct evaluation dimensions,
each with specialized prompts for precise assessment.

The task of T2V differs from I2V, as videos generated from the same text can vary widely, making it less suitable for evaluation metrics that require a reference video. For T2V tasks, the results generated by different models for the same text prompt can be quite dissimilar. However, for I2V tasks, since the image imposes certain constraints, the variation in results produced by different models is generally not as pronounced. This allows us to conduct a comprehensive evaluation of different Image-to-Video (I2V) algorithms on video-text datasets using evaluation metrics that are based on reference videos. Our AIGCBench draws on these T2V benchmarks but differs from them in several respects:
1). We need to collect or construct images for the I2V model's input, which requires considering the comprehensiveness of both the text prompt set and the image set.
2). Although our evaluations are similar to those of the T2V task in terms of the dimensions assessed, we need to employ new evaluation standards due to the differences between T2V and I2V tasks. 


\subsection{Benchmarks for Image-to-Video Generation}
\label{sec:i2vbench_related}

\paragraph{Domain-specific I2V benchmark.}
LFDM Eval~\cite{ni2023conditional} is evaluated on facial expression and human action datasets, employing just a few evaluation metrics to gauge the quality of video generation.
The CATER-GEN~\cite{hu2023benchmark} benchmark uses predefined 3D objects and specific initial images for testing the quality of videos that depict the motion of 3D objects.
Nonetheless, neither LFDM Eval~\cite{ni2023conditional} nor the CATER-GEN~\cite{hu2023benchmark} benchmark is appropriate for evaluating video generation in open-domain scenarios.

\paragraph{Open-domain I2V benchmark.}

The open-domain I2V benchmark is currently based on two main types of evaluation data: video-text and image-text datasets.
Seer~\cite{gu2023seer} and SVD~\cite{blattmann2023stable} have utilized video-text datasets and employed a limited number of metrics that require reference videos for evaluation.
VideoCrafter~\cite{chen2023videocrafter1} and I2VGen-XL~\cite{zhang2023i2vgen} have used image-text datasets and relied solely on visual comparisons. Very recently, AnimateBench~\cite{zhang2023pia} was released for the purpose of evaluating I2V tasks. They also generated images using Text-to-Image models. However, they were limited by a small number of text prompts and a limited collection of images. At the same time, there is a lack of comprehensive evaluation metrics. Both are constrained by limited evaluation datasets and an incomplete set of assessment metrics. This leads to the evaluation datasets not being representative of all stakeholders' concerns and interests, and there is also a lack of a unified and comprehensive consensus on evaluation. In this paper, we expand the image-text dataset using state-of-the-art Text-to-Image models. 
To ensure the complexity of the generated text prompts, we generate prompts through the combinatorial traversal of four metatypes and enhance them with the capabilities of large language models. 
We compare our AIGCBench with other I2V benchmarks in Table~\ref{tab:compare_benchmark}.

\paragraph{Generating Image-Text Dataset} While most benchmarks gather datasets from the real world, CATER-GEN~\cite{hu2023benchmark} constructs datasets using a limited set of text prompts for specific object movement scenarios. Very recently, AnimateBench~\cite{zhang2023pia} utilized a limited number of manually designed text prompts and also employed Text-to-Image models to generate images. However, this approach is constrained by the simplicity of the text combinations and the limited diversity of the images. Our generation pipeline uses a text combiner to randomly generate text prompts and incorporates GPT-4~\cite{openai2023gpt4} to enrich the content. Simultaneously, we filter the generated results to select high-quality image-text pairs.


\section{AIGCBench: Establishing the Image-to-Video Generation Benchmark}
\label{sec:benchmark}

The framework of our AIGCBench is shown in Figure~\ref{fig:framework}. Our AIGCBench framework comprises three components: the evaluation dataset, the video generation models to be assessed, and the evaluation metrics. To construct a comprehensive benchmark, we evaluate I2V models using two types of datasets: video-text and image-text. For the image-text dataset, we utilize evaluation metrics that do not require reference videos. In this section, we will introduce how we collected the evaluation datasets, in Section~\ref{sec:evaluation} we present the evaluation criteria we have established, and in Section~\ref{sec:exp_models} we provide a brief introduction to the video generation models to be evaluated.

\begin{figure*}
    \centering
    \includegraphics[scale=.52]{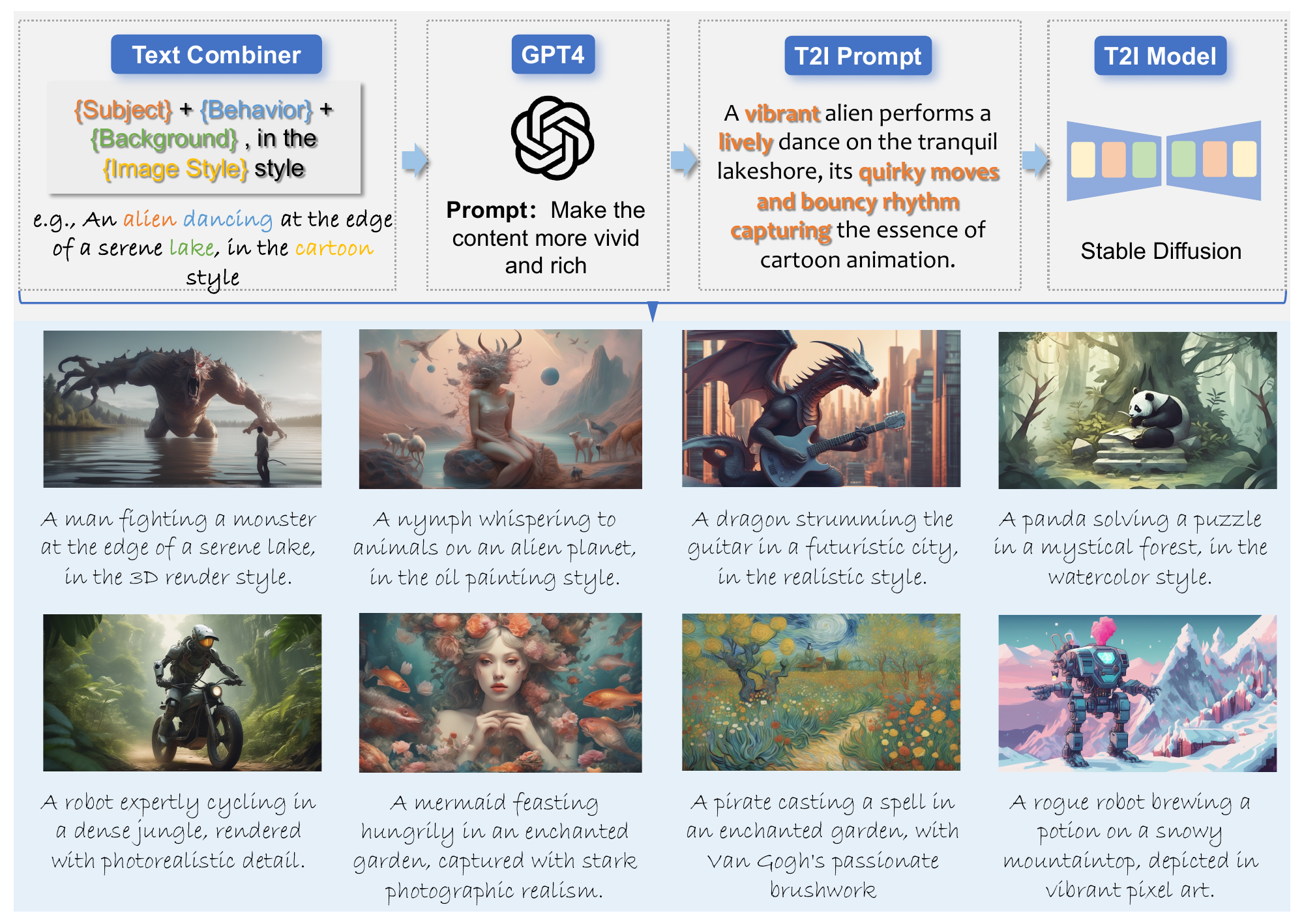}
    \caption{Image-text dataset generation pipeline and results. Above: An overview of our T2I generation pipeline is presented. Below: Eight generated cases are showcased, with the original text produced by the text combiner displayed beneath each image.}
    \label{fig:generation_pipeline}
\end{figure*}

\subsection{Collect Dataset from Real-World}

\paragraph{Video-Text Pairs} The WebVid-10M~\cite{bain2021frozen} dataset is a substantial collection specifically designed to aid in the development and training of AI models for video understanding tasks. It consists of approximately 10 million video-text pairs, making it one of the larger datasets available for this type of research. Considering that video generation is time-consuming, we have sampled 1,000 videos from the validation set of the WebVid10M~\cite{bain2021frozen} dataset based on subtype for evaluation purposes.

\paragraph{Image-Text Pairs} The LAION-5B~\cite{schuhmann2022laion5b} dataset is a large-scale, open dataset consisting of around 5,85 billion image-text pairs. It was created to facilitate research in computer vision and machine learning, specifically in areas such as multi-modal language-vision models, Text-to-Image generation, and more(e.g. CLIP~\cite{radford2021learning}, DALL-E~\cite{ramesh2021zero}). LAION-Aesthetics is a subset from LAION-5B~\cite{schuhmann2022laion5b} with high visual quality. We randomly sampled 925 image-text pairs from the LAION-Aesthetics dataset to serve as a reference for video-free evaluation metrics.


\subsection{Generated Image-Text Pairs}

Using only real-world datasets is insufficient. Users often input images and text generated by designers or T2I (Text-to-Image) models to create videos. This includes certain image-text pairs that cannot be sampled in the real world. To bridge this gap, we propose a T2I generation pipeline. As shown in Figure~\ref{fig:generation_pipeline}, we provide an overview of our generation pipeline above and present some generated cases below.

\subsubsection{Text Combiner}
To generate as diverse text prompts as possible, we construct text templates based on four types: subject, behavior, background, and image style. We then generate a list of 3,000 text prompts randomly by following the template: {\textbf{subject}} + {\textbf{behavior}} + {\textbf{background}}, in the {\textbf{image style}} style.
We have listed some examples:
\begin{enumerate}
\itemsep=0pt
\item Subject: \textsl{a dragon, a knight, an alien, a robot, a panda, a nymph};
\item Behavior: \textsl{riding a bike, fight a monster, searching for a treasure, dancing, solving a puzzle};
\item Background: \textsl{in a forest, in a futuristic city, in a space station, in an old western town at high noon};
\item Image style: \textsl{oil painting, water color, cartoon, realistic, Van Gogh, Picasso}.

\end{enumerate}  


We have compiled our text corpus from high-frequency words often entered by users in the T2I community of Civit AI~\cite{civitai}, along with some potentially valuable text prompts. Considering the flexibility of our generation pipeline, our benchmark is scalable. Subsequently, we can update and iterate on the versions of our text corpus.

\subsubsection{Optimizing text prompts}

Although utilizing text templates with various text corpora can generate reasonable images, it might lead to poor diversity in the generated images, which is not conducive to evaluating I2V tasks. We leverage the capabilities of the GPT-4 model~\cite{openai2023gpt4}, using the prompt "\textsl{make the content more vivid and rich}" to optimize the texts generated from templates.

\subsubsection{Generate images and filter}

To generate high-quality images based on the generated texts, we have employed the best Text-to-Image (T2I) model available to date – the Stable Diffusion model~\cite{rombach2022high}.
The Stable Diffusion model~\cite{rombach2022high} is particularly notable for its ability to create high-quality and coherent images that closely match the style and content described by the input text prompts. We utilized the latest xl-base T2I model released by their community. Considering that the I2V model is primarily trained with an aspect ratio of 16:9, we used a height of 720 and a width of 1280 to generate images.

In order to select high-quality image-text pairs, we filtered out the top 2003 high-quality image-text pairs based on the automatic metrics from the T2I-CompBench~\cite{huang2023t2i}.
Some examples generated by our pipeline can be seen in the lower half of Figure~\ref{fig:generation_pipeline}.

\section{Evaluation Metrics}
\label{sec:evaluation}
Our evaluation dataset includes both video-text and image-text datasets. To conduct a comprehensive evaluation, we employ two types of assessment metrics: one that requires reference videos and another that does not. In addition, we considered previous Text-to-Video benchmarks~\cite{liu2023evalcrafter,liu2023fetv,huang2023vbench} and have integrated to propose an evaluation standard suitable for the Image-to-Video (I2V) task, covering both types of dataset. We assess the performance of different I2V models from four aspects: control-video alignment\footnote{"control" refers to the input signals from the user, such as text, images, and other forms of control signals.}, motion effects, temporal consistency, and overall video quality\footnote{The code is available at \href{https://github.com/BenchCouncil/AIGCBench}{https://github.com/BenchCouncil/AIGCBench}.}.
Considering that videos generated by different algorithms have varying numbers of frames, for a standardized evaluation, we adopt the approach of extracting the first 16 frames, unless otherwise specified.

\subsection{Control-video alignment}

The control-video alignment measures the degree of alignment between the user's input control signals, such as text and images, and the generated video. Considering that current video generation tasks primarily involve two types of inputs—a starting image and a text prompt—we introduce two evaluation metrics in the first version of our benchmark: \textbf{image fidelity} and \textbf{text-video alignment}.
The image fidelity metric evaluates how similar the generated video frames are to the image input into the I2V model, especially the first frame. To assess fidelity, for the first frame of the generated video, we use metrics such as Mean Squared Error~(MSE) and Structural Similarity Index Measure (SSIM)~\cite{wang2004image} to calculate the degree of preservation of the first frame. For all frames of the video, we calculate the similarity of CLIP (Contrastive Language–Image Pre-training)~\cite{radford2021learning} embeddings between the input image and each frame of the generated video.
We use MSE (First), SSIM (First), and Image-GenVideo CLIP to represent these three evaluation metrics, respectively.


Considering that the I2V models we evaluate also take text as input, we need to assess whether the generated videos are relevant to the input text. For the generated videos, we use CLIP~\cite{radford2021learning} to calculate the similarity between the input text and the generated video results. We assume that the videos in the video-text dataset are consistent with the textual descriptions. For the video-text dataset, we use the keyframes from the reference videos and the generated videos to compute the CLIP~\cite{radford2021learning} similarity. Considering that the text typically describes high-level semantics and that the generated videos may not correspond perfectly with the original videos, we uniformly sample four keyframes for comparison. We use GenVideo-Text Clip and GenVideo-RefVideo CLIP (Keyframes) to represent these two evaluation metrics, respectively.

\subsection{Motion effects}
\label{sec:mq}
Motion effects primarily evaluate whether the amplitude of the motion in the generated video is significant and whether the movements are reasonable.
As for the amplitude of the motion, we follow the~\cite{liu2023evalcrafter,huang2023vbench} and use a pretrained optical flow estimation method, RAFT~\cite{teed2020raft}, to calculate the flow score between adjacent frames of the generated video, with the final average value representing the magnitude of the motion effects. We use the square average of the predicted values from adjacent frames to represent the motion dynamics of the video, with higher values indicating stronger motion effects.
Considering that there are some bad cases in video generation, we set a threshold where the square average value must be less than 10 to filter out these bad cases.
For the video-text dataset, we have real videos corresponding to the text. We measure the reasonableness of the generated motion effects by calculating the similarity between each frame of the generated video and each frame of the reference video, and then taking the average. For robustness, we use the image CLIP~\cite{radford2021learning} metric to calculate the similarity between frames. 
We use Flow-Square-Mean and GenVideo-RefVideo CLIP (Corresponding frames)  to represent these two evaluation metrics, respectively.



\subsection{Temporal Consistency}

Temporal consistency measures whether the generated video frames are consistent and coherent with each other. We calculate the image CLIP~\cite{radford2021learning} similarity between every two adjacent frames in the generated video and take the average as an indicator of the temporal consistency of the generated video. We use GenVideo Clip (Adjacent frames) to represent this evaluation metric. In addition, we also use GenVideo-RefVideo (Corresponding frames) from Section \ref{sec:mq} to represent temporal consistency.

\subsection{Video Quality}

Video quality is a relatively subjective dimension, measuring the overall quality of video production. We first use the number of frames generated by videos to gauge the ability of different algorithms to generate long videos. We utilize disentangled
objective video quality evaluator (DOVER)~\cite{wu2023exploring}, a no-reference video quality assessment metric. DOVER~\cite{wu2023exploring} comprehensively rates videos from both aesthetic and technical perspectives, using the collected DIVIDE-3k dataset. Experimental results show that the DOVER~\cite{wu2023exploring} metric highly correlates with human opinions in both aesthetic and technical perspectives. For the DOVER evaluation metric, we calculate it using all frames generated by their respective algorithms. For the video-text dataset, since we have reference videos available, we measure the spatial structural similarity of the generated videos to the reference videos by calculating the SSIM (Structural Similarity Index Measure) between the corresponding frames of the generated and reference videos. We denote this evaluation metric as GenVideo-RefVideo SSIM.

\begin{table*}[t]
\centering
\resizebox{\textwidth}{!}{%
\begin{tabular}{c|l|c|c|c|c|c}
\toprule
Dimensions   & Metrics   & VideoCrafter~\cite{chen2023videocrafter1} & I2VGen-XL~\cite{zhang2023i2vgen} & SVD~\cite{blattmann2023stable} & Pika~\cite{discordpika} & Gen2~\cite{esser2023structure}  \\ \hline 
\multirow{5}{*}{\shortstack{Control-video\\Alignment}}
& MSE (First) $\downarrow$  & 3929.65         & 4491.90     & 640.75            & \textbf{155.30}   & 235.53 \\ \cline{2-7} 
& SSIM (First)     $\uparrow$   & 0.300          & 0.354       & 0.612              & 0.800    & \textbf{0.803} \\ \cline{2-7} 
& Image-GenVideo Clip       $\uparrow$ & 0.830   & 0.832     & 0.919          & 0.930   & \textbf{0.939} \\ \cline{2-7}
& GenVideo-Text Clip   $\uparrow$    & 0.23  & 0.24  &     -     & \textbf{0.271}   & 0.270  \\ \cline{2-7} 
& GenVideo-RefVideo CliP (Keyframes)  $\uparrow$       & 0.763      &  0.764       &     -     & \textbf{0.824}        &    0.820   \\ \hline
\multirow{2}{*}{\shortstack{Motion\\Effects}}
& Flow-Square-Mean  & 1.24         & 1.80    & 2.52            & 0.281   & 1.18 \\ \cline{2-7} 
& GenVideo-RefVideo CliP (Corresponding frames)     $\uparrow$   & 0.764          & 0.764       & 0.796              & \textbf{0.823}    & 0.818 \\ \hline
\multirow{2}{*}{\shortstack{Temporal\\Consistency}} & GenVideo Clip (Adjacent frames)  $\uparrow$   & 0.980  & 0.971     & 0.974    & \textbf{0.996}   & 0.995 \\ \cline{2-7} 
& GenVideo-RefVideo CliP (Corresponding frames)     $\uparrow$   & 0.764          & 0.764       & 0.796              & \textbf{0.823}    & 0.818 \\ \hline
\multirow{3}{*}{\shortstack{Video\\Quality}}
& Frame Count $\uparrow$  & 16         & 32     & 25            & 72   & \textbf{96} \\ \cline{2-7} 
& DOVER     $\uparrow$   & 0.518          & 0.510       & 0.623              & 0.715    & \textbf{0.775} \\ \cline{2-7} 
& GenVideo-RefVideo SSIM       $\uparrow$ & 0.367   & 0.304     &    0.507       & \textbf{0.560}   & 0.504 \\ \bottomrule
\end{tabular}}
\caption{
Quantitative analysis for different Image-to-Video algorithms. An upward arrow indicates that higher values are better, while a downward arrow means lower values are preferable.
}
\label{tab:results}
\end{table*}

\section{Experiments}
\label{sec:experiments}

\subsection{Evaluated models}

\label{sec:exp_models}

\subsubsection{Open-source project}
\paragraph{VideoCrafter} VideoCrafter~\cite{chen2023videocrafter1} is an open-source video generation and editing toolbox for crafting video content. It supports the generation of videos from images. We use a guidance scale of 12 and ddim steps of 25. For videos with an aspect ratio of 1, we employ a resolution of 512 * 512, while for videos with an aspect ratio of 0.5625, we use a resolution of 512 * 320, and then uniformly resize to align with the resolutions used by other methods.

\paragraph{I2VGen-XL} I2VGen-XL~\cite{zhang2023i2vgen} is an open-source video synthesis codebase developed by Tongyi Lab at Alibaba Group, which features state-of-the-art video generative models. We use a guide scale of 9 and infer with fp16 precision.

\paragraph{Stable Video Diffusion} Stable Video Diffusion (SVD)~\cite{blattmann2023stable} is an expansion of the model based on Image Stable Diffusion~\cite{rombach2022high}. We use the 25-frame version of Stable Video Diffusion. It is worth noting that the current model does not support text input temporarily, hence we did not calculate the text-video alignment for this model.

\subsubsection{Closed-source project}

\paragraph{Pika} Pika~\cite{discordpika} is a technology company revolutionizing video creation by making it effortless and accessible for everyone. In just six months, Pika has built a community of half a million users producing millions of videos per week. The company recently launched Pika 1.0, a significant upgrade featuring a new AI model that supports various video styles, including 3D animation, anime, cartoons, and cinematic, coupled with an improved web experience. Considering that Pika~\cite{discordpika} does not have open-source code, we manually tested 60 cases on the Discord platform (30 from the WebVid dataset and 30 from our own generated dataset). We used the default parameters of motion set to 1 and the guidance scale set to 12.

\paragraph{Gen2} Gen2~\cite{esser2023structure} is a multimodal AI system that can generate novel videos with text, images, or video clips. We used the default motion setting of 5 from the demo and did not employ the camera movement parameter to generate videos.

\begin{figure*}
    \centering
    \includegraphics[scale=.485]{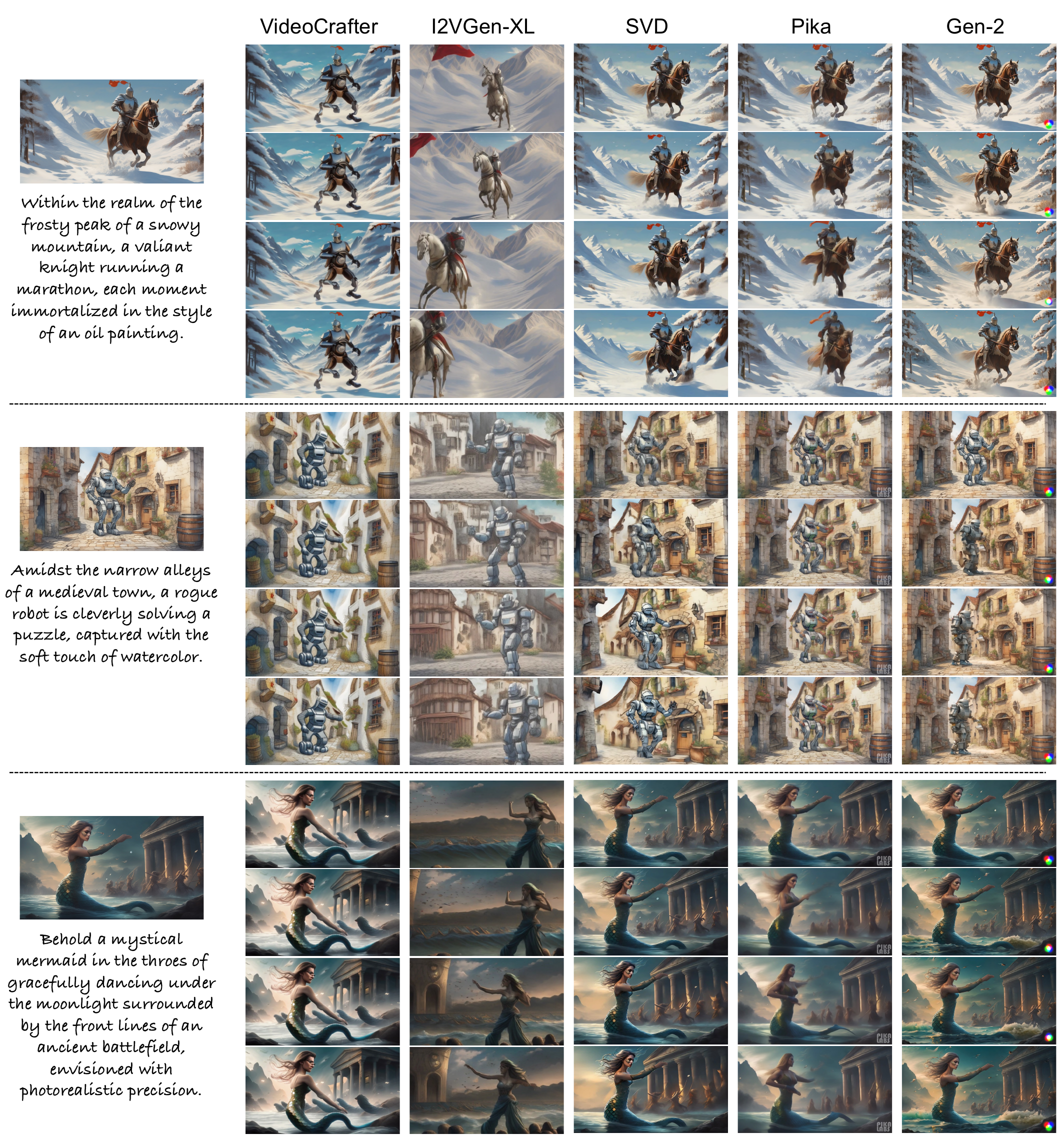}
    \caption{We present three I2V cases utilizing five state-of-the-art algorithms, among which VideoCrafter, I2VGen-XL, and SVD are open-source research, while Pika and Gen2 are closed-source project. For additional videos, please refer to our project website.}
    \label{fig:comparemethod}
\end{figure*}

\subsection{Comprehensive Results Analysis}

Table~\ref{tab:results} presents the evaluation of five state-of-the-art (SOTA) I2V algorithms across five dimensions: image fidelity, motion effects, text-video alignment, temporal consistency, and video quality. We present the qualitative results of different I2V algorithms in Figure~\ref{fig:comparemethod}. We find that VideoCrafter and I2VGen-xl struggle to preserve the original image. I2VGen-xl maintains relatively good semantics, but the spatial structure of the initial image is mostly not preserved. VideoCrafter can approximate the spatial structure of the initial image to some extent, but the preservation of details is generally mediocre. SVD, Pika, and Gen2 preserve the original image quite well, with Gen2 achieving the best preservation effect. As for the aspect of Text-video alignment, Gen2 and Pika are nearly on par with each other and both outperform the open-source algorithms. However, existing algorithms and evaluation metrics do not effectively capture fine-grained textual changes. In terms of motion effects, VideoCrafter tends to remain static. I2VGen-xl and SVD lean towards camera movement rather than subject motion, which is why they score high on the flow-square-mean but obtain low GenVideo-RefVideo Clip scores. Pika tends to favor both local and subject movement, thus achieving high GenVideo-RefVideo Clip scores and low flow-square-mean scores. Gen2, on the other hand, favors movement in both the foreground and background, but the background movement is not as pronounced as with SVD. 

In the aspect of temporal Consistency, VideoCrafter, due to its poorer motion effects, does not perform poorly in terms of temporal consistency. Considering that SVD has stronger motion effects and still maintains good temporal consistency, it has achieved the best performance among open-source I2V algorithms. Similarly, Pika, because of its tendency for local movement, has achieved the highest score in overall temporal consistency. 
As for video quality, Gen2 is capable of generating the longest videos of up to 96 frames, with the highest levels of aesthetics and clarity. Pika, due to its tendency for local movement, has achieved the highest similarity in the GenVideo-RefVideo SSIM metric. SVD benefits from the priors of the image stable diffusion model, resulting in videos that reach the best performance among open-source I2V algorithms. In summary, the two closed-source projects, Pika and Gen2, achieved the most optimal generation effects, capable of producing long videos. Pika excels in generating local motion, while Gen2 tends to prefer global motion. SVD achieved the best results among the open-source options, demonstrating outcomes that were close to those of the two closed-source projects. 


\begin{figure}
    \centering
    \includegraphics[scale=.37]{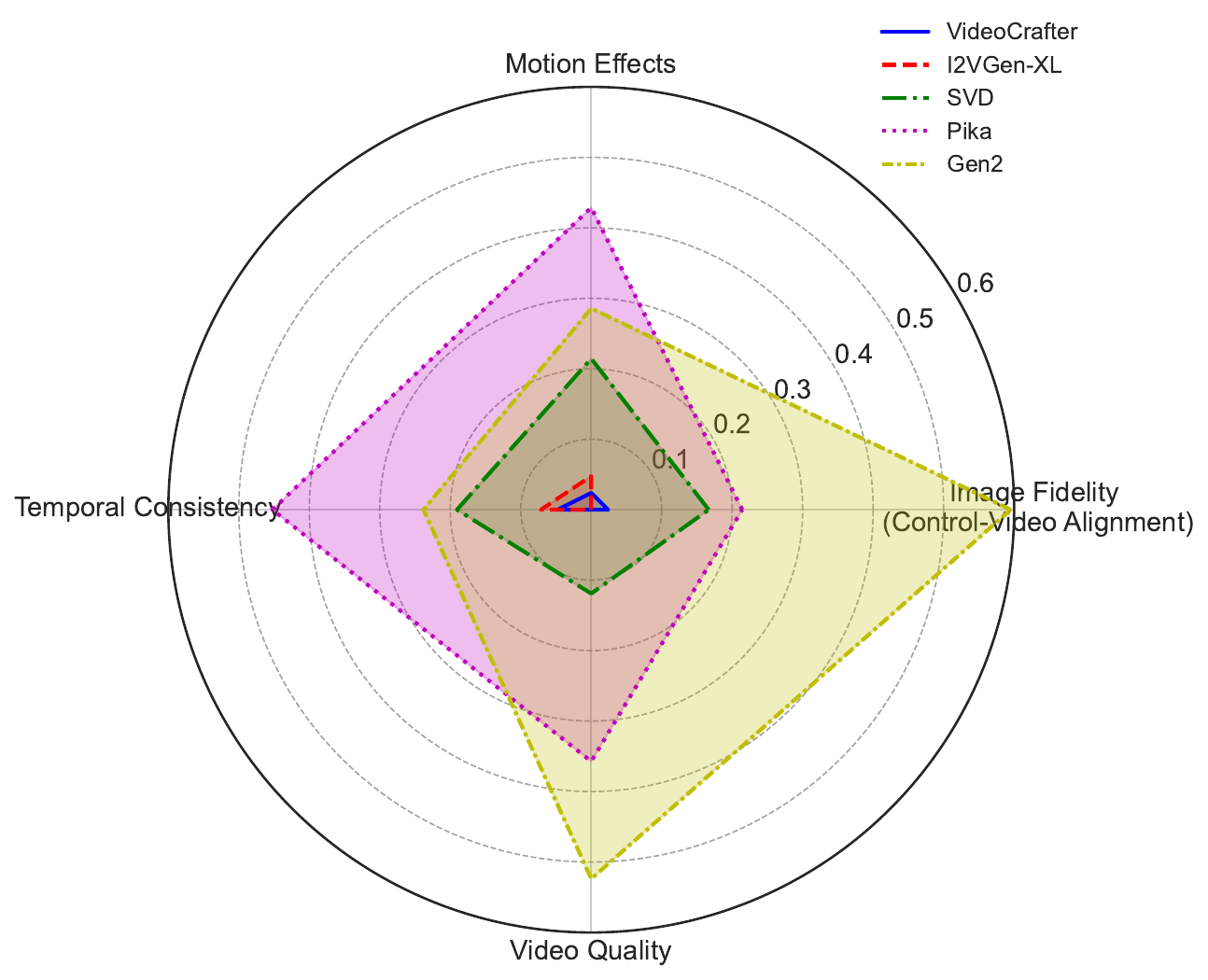}
    \caption{We tallied the votes of 42 individuals, evaluating five state-of-the-art I2V algorithms from four aspects. The numerical values in the radar chart represent the proportion of users who voted for each algorithm as being the best performer in that aspect.}
    \label{fig:radar}
\end{figure}

\subsection{User study}

To validate whether the proposed evaluation standards are aligned with human preference, we randomly sampled 30 generated results from each of the five methods and tallied the best algorithm outcomes in each of the four dimensions (Image Fidelity, Motion Effects, Temporal Consistency, Video Quality) through human voting. We have tallied the votes of a total of 42 individuals, with the specific results presented in Figure~\ref{fig:radar}. We discovered that Gen2's performance is on par with Pika, both achieving optimal results. Pika excelled in temporal consistency and motion effects, while Gen2 came out on top in terms of image fidelity and video quality. SVD showed a balanced performance across all areas, securing the best results among the open-source options. We found that the users' votes are relatively consistent with the results evaluated by our assessment criteria. 

\subsection{Findings and Discussions}
\label{sec:discussions}

Despite the notable achievements of I2V and the rapid updates of new algorithms, there is still significant room for improvement in existing solutions. Utilizing AIGCBench, we have conducted a detailed survey and evaluation of the five most advanced I2V algorithms from both academia and industry. The comprehensive analysis facilitated by AIGCBench has led us to make the following discoveries:

\paragraph{Lacking fine-grained control} AIGCBench's testing methodology has revealed that current I2V tasks often fall short in allowing users to generate content with precise textual descriptions. The benchmark's diverse datasets and nuanced evaluation metrics have shown that while solutions based on CLIP~\cite{radford2021learning} and large language models~\cite{raffel2020exploring} are a step in the right direction, they are not fully capturing the fine-grained details that yield a high degree of control over video content. AIGCBench's detailed feedback on these algorithms has underscored the need for a model that is specifically trained for video contexts to improve text-video alignment and to provide users with that control. Our findings suggest that integrating AIGCBench's evaluation criteria into the development process could lead to algorithms that better align with human preferences.

\paragraph{Long video generation} The current I2V algorithms can generate up to 96 frames in a single inference, which is far from satisfying users' needs for longer video production. Considering that video scenes typically have a frame rate of 24 fps, the basic generation capability of mainstream algorithms is around 3 seconds. There are mainly two approaches to address this limitation. One is to use multiple inferences, where most adopt a coarse-to-fine generation pipeline—first generating diluted keyframes, then densely producing all frames. The challenge of this method lies in maintaining temporal consistency across multiple inferences. The other approach is to use multi-GPU training and inference with a single model, which currently struggles to guarantee satisfactory results. How to generate longer videos should be an urgent issue for the AI-generated content (AIGC) community to address next.

\paragraph{Inference speed} Currently, the speed of video generation is relatively slow. For a 3-second video, mainstream algorithms generally require about 1 minute on a V100 graphics card. Considering that video generation scenarios are based on diffusion models~\cite{sohl2015deep,song2019generative,ho2020denoising,nichol2021improved,dhariwal2021diffusion}, there are currently two main routes for speeding up the process. One is to reduce the dimensionality of the video in the latent space. For example, Stable Diffusion~\cite{blattmann2023stable} maps the video into a latent space, roughly decreasing the size of the video by about 8 times, with only a minimal loss of video quality. The other is to improve the inference speed of the diffusion model, which is also a hot research topic in the AIGC community.

In light of these findings from AIGCBench, it becomes evident that the benchmark not only aids in the identification of current shortcomings but also offers a structure for addressing them. AIGCBench’s comprehensive framework for evaluation, encompassing a rich variety of datasets and multi-dimensional metrics, provides a roadmap for advancing the state of I2V algorithms. By exposing algorithmic weaknesses and offering a standardized platform for comparison, AIGCBench guides researchers towards developing solutions that overcome the challenges of fine-grained control, video length, and inference speed. The insights gained from AIGCBench evaluations are instrumental in pushing the boundaries of what is possible in the field of I2V generation.

\section{Conclusion}



In this work, we have introduced AIGCBench, a comprehensive and scalable benchmark tailored for the evaluation of Image-to-Video (I2V) generation tasks. AIGCBench provides a much-needed framework to assess the performance of various state-of-the-art I2V algorithms under equivalent evaluation conditions. Our benchmark stands out by incorporating a diverse set of real-world video-text and image-text datasets, as well as a novel dataset produced through our proprietary generation pipeline.
We have also proposed a novel set of evaluation metrics that span across four critical dimensions: control-video alignment, motion effects, temporal consistency, and video quality. These metrics have been validated against human judgment to ensure their alignment with human preferences. Our extensive evaluation of leading I2V models has not only highlighted their strengths and weaknesses but also unearthed significant insights that will guide the future development of the I2V domain.

AIGCBench marks a foundational step in benchmarking for AIGC, pushing the frontier of I2V technology evaluation. By offering a scalable and precise assessment methodology, we set the stage for continuous enhancements and innovations in this rapidly evolving research field. As we progress, we plan to expand AIGCBench to encompass a broader range of video generation tasks, creating a unified and extensive benchmark that reflects the multifaceted nature of AIGC.

\section{Limitations and Future Work}

Due to the slow inference speed of video generation by I2V models and the fact that some works are not open-sourced (e.g., Pika~\cite{discordpika}, Gen2~\cite{esser2023structure}), our benchmark only evaluated 3950 test cases. Considering the complexity of video generation tasks, we believe this number is insufficient. Furthermore, given the lack of fine-grained video recognition models currently available, our evaluation system is unable to accurately judge whether the direction of object movement in the generated videos matches the text description. For instance, whether water flows from left to right or from right to left, we are currently unable to determine through automated evaluation metrics if the direction of the water flow in the generated video is consistent with the textual description.

Moving forward, we will integrate tasks related to T2V and new video generation tasks into a large-scale video generation benchmark. Additionally, to address the issues mentioned above, we may train a fine-grained video representation model aligned with text, which will be utilized for fine-grained alignment of video and text scenes.

\section*{Acknowledgments}

I would like to extend my heartfelt thanks to Professor Lei Wang for his insightful discussions and valuable revisions to this manuscript. I am also grateful to Mengya He for her contributions to the discussions of this paper. Furthermore, I wish to acknowledge Litong Gong, Weijie Li, Yiran Zhu, and Biao Wang from Alibaba Company for their support in the experimental aspects of this research.

\printcredits

\bibliographystyle{cas-model2-names}

\bibliography{AIGCBench}





\end{sloppypar}
\end{document}